\documentclass[runningheads]{llncs}
\usepackage[T1]{fontenc}
\usepackage{graphicx}
\usepackage{xcolor}
\usepackage{cite}
\usepackage{amsmath,amssymb,amsfonts}
\usepackage{mathtools}
\usepackage{algorithmic}
\usepackage{graphicx}
\usepackage{textcomp}
\usepackage{mathtools}
\usepackage{multicol, blindtext}  
\usepackage{bm}

\usepackage{booktabs} 

\usepackage{algorithmic}

\newtheorem{assumption}{Assumption}

\begin{document}
\title{Continuous-Time Robust Control for Cancer Treatment Robots
}
%
%
\author{Vlad Mihaly\inst{1}
\and Iosif Birlescu\inst{2}
\and Mircea Șușcă\inst{1}
\and Damien Chablat\inst{2,3}
\and Petru Dobra\inst{1}}
\authorrunning{V. Mihaly et al.}
%
\institute{Department of Automation, Technical University of Cluj-Napoca, Romania \and
CESTER, Technical University of Cluj-Napoca, Romania \and
École Centrale Nantes, Nantes Université, CNRS, LS2N, Nantes, France \\
Corresponding author: \email{ iosif.birlescu@mep.utcluj.ro}
}
\maketitle              
\vspace{-0.5cm}
\begin{abstract}
The control system in surgical robots must ensure patient safety and real time control. As such, all the uncertainties which could appear should be considered into an extended model of the plant. After such an uncertain plant is formed, an adequate controller which ensures a minimum set of performances for each situation should be computed. As such, the continuous-time robust control paradigm is suitable for such scenarios. However, the problem is generally solved only for linear and time invariant plants. The main focus of the current paper is to include m-link serial surgical robots into Robust Control Framework by considering all nonlinearities as uncertainties. Moreover, the paper studies an incipient problem of numerical implementation of such control structures. 

\end{abstract}
\section{Introduction}
Robotic-assisted cancer treatment was introduced in the 20th century, showing distinct advantages over classical interventions (whether we refer to surgical or percutaneous interventions), such as better accuracy, better ergonomics, and safety \cite{Haidegger:22,EU:22,Balaur:09,Balaur:19}. The cancer treatment robotics field is still in continuous development, with more state-of-the-art and emerging technologies being implemented into the surgical robotic systems such as Artificial Intelligence (AI), advanced vision, smart safety features and control modules \cite{Haidegger:22}. One trend in the emerging cancer treatment robotic systems is to provide better surgical outcomes through personalized instrumentation and intervention\cite{EU:22}.

Continuous-time robust control synthesis has found extensive applications in various practical scenarios due to its high flexibility, and should be suitable for implementation in cancer treatment robotic systems. This approach offers a versatile framework for extending nominal models described by transfer matrix or state-space representations with model uncertainties and performance specifications through open-loop \cite{McFarlane:92} and closed-loop shaping \cite{Skogestad:05}. To design unstructured regulators, two approaches are the most common: algebraic Riccati equations \cite{Zhou:95} and linear matrix inequalities \cite{Liu:16}. However, these controllers have the same order as the plant, leading to implementation issues. To design fixed-structure controllers, a nonsmooth optimization method has been presented in \cite{Apkarian:17}. Commonly used performance metrics are the $\mathcal{H}_2$ and $\mathcal{H}_{\infty}$ norms, employed to quantify system performance, both in continuous and discrete cases \cite{Tudor:15}. The $\mathcal{H}_2/\mathcal{H}_{\infty}$ norms have been further extended by incorporating the structured singular value (SSV), which efficiently captures uncertainty in plant dynamics \cite{Apkarian:11}.

This framework has been used in \cite{mihaly2022ECC} as an extra layer to guarantee the robustness of a nonlinear systems with polytopic approximation. The m-link serial robots studied in this paper have polytopic approximation. The main contribution of the current paper consists in removing the inner layer of the control structure proposed in \cite{mihaly2022ECC}, to simplify the design. The matrices from the polytopic approximation will be considered as uncertainties against a nominal plant obtained at the equilibrium point. As such, the contributions of the paper are: (1) obtaining a polytopic differential inclusion representation of an m-link serial robot; (2) representing an m-link serial robot (as a generic robot for cancer treatment) using polytopic differential inclusions; (3) including the nonlinearities into additive uncertainties against a linearized model around a given equilibrium point; (4) applying the generalized Robust Control Framework on the given problem; (5) illustrating the proposed control structure on a 2R serial robot (illustrating the possibility of integration in surgical robots).

The rest of the paper is organized a s follows: Section 2 presents the problem to be solved, along with the available solutions and how to adapt them for our problem; in Section 3 the case of 2R serial robot is presented into an end-to-end manner, while Section 4 presents conclusions and further research directions. 

\textit{Notations:} $\text{Co}\left(S\right)$ is the convex hull of the set $S$. The sets of symmetric and positive-(semi)definite matrices of order $m$ are $\mathbb{S}_{m}^{\geq 0}$ and $\mathbb{S}_m^{+}$. The variable $s$ is the complex frequency used in the Laplace Transform.

\section{Problem Formulation}

\begin{figure}[!b]
    \centering
    \includegraphics[width=0.45\columnwidth]{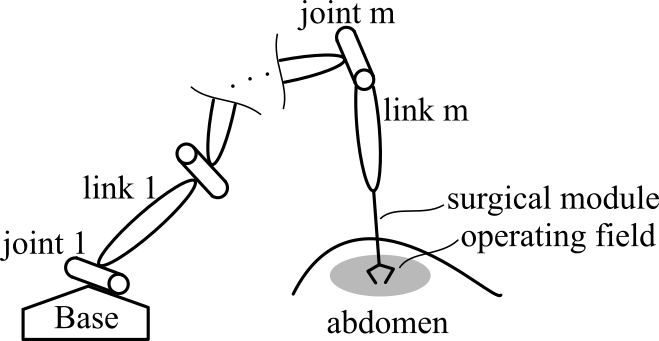}
    \caption{General m-link cancer treatment serial robot.}
    \label{fig:m_link_fig}
\end{figure}

Consider a general m-link serial robot (that manipulates a surgical instrument for cancer treatment) as in Figure \ref{fig:m_link_fig} described by the following dynamic model: 
\begin{equation}
\label{eq:m_link_robot}
    M(\mathbf{q})\ddot{\mathbf{q}} + C(\mathbf{q},\dot{\mathbf{q}})\dot{\mathbf{q}} + D\dot{\mathbf{q}} + g(\mathbf{q}) = \mathbf{u},
\end{equation}
where $\mathbf{q}\in \mathcal{D}_{q}\subset \mathbb{R}^m$ is the vector of generalized coordinates representing joint positions,
$\mathbf{u}\in \mathbb{R}^m$ is the control input vector, while $M(\mathbf{q})$ is the inertia matrix, $C(\mathbf{q}, \dot{\mathbf{q}})\dot{\mathbf{q}}$ encompasses the centrifugal and Coriolis forces, $D\dot{\mathbf{q}}$ is the viscous damping term, and $g(\mathbf{q})$ is the gravity term. The gravity term is given by: 
\begin{equation}
    g(\mathbf{q}) = \left(\frac{\partial P(\mathbf{q})}{\partial \mathbf{q}}\right)^{\top},
\end{equation}
where $P(\mathbf{q})$ is the total potential energy of the links due to gravity.

\begin{assumption}
    For each $\mathbf{q}\in\mathcal{D}_q$ we assume that the following conditions are satisfied:  $M(\mathbf{q})\in \mathbb{S}_{m}^{+}$, $\dot{M}-2C$ is skew-symmetric, and $D\in\mathbb{S}_{m}^{\geq 0}$.
\end{assumption}

The state vector $\mathbf{x}\in\mathcal{D}_x\subset \mathbb{R}^{n_x}$ contains the generalized coordinates $\mathbf{x}_1 = \mathbf{q}$ and velocities $\mathbf{x}_2 = \dot{\mathbf{q}}$, so $n_x = 2\cdot m$. The state-space model is given by: 
\begin{equation}
\label{eq:m_link_robot_ss}
    \begin{cases}
        \dot{\mathbf{x}}_1 = \mathbf{x}_2; \\
        \dot{\mathbf{x}}_2 = -M^{-1}(\mathbf{x}_1)\left(C(\mathbf{x}_1,\mathbf{x}_2)+D\right)\mathbf{x}_{2} - M^{-1}(\mathbf{x}_1)g(\mathbf{x}_1) + M^{-1}(\mathbf{x}_1)\mathbf{u},
    \end{cases}
\end{equation}
which could be viewed as an input-affine nonlinear system \cite{Khalil}:
\begin{equation}
    (\Sigma): \dot{\mathbf{x}} = f_0(\mathbf{x}) + \sum_{i=1}^m f_i(\mathbf{x})u_i \equiv f(\mathbf{x},\mathbf{u}),
\end{equation}
where: 
\begin{subequations}
\begin{equation}
    f_0(\mathbf{x}) = \begin{pmatrix}
        \mathbf{x}_2 \\ -M^{-1}(\mathbf{x}_1)\left(C(\mathbf{x}_1,\mathbf{x}_2)+D\right)\mathbf{x}_{2} - M^{-1}(\mathbf{x}_1)g(\mathbf{x}_1)
    \end{pmatrix};
\end{equation}
\begin{equation}
    \mathbf{f}(\mathbf{x}) = \begin{pmatrix}
        f_1(\mathbf{x}) & \dots & f_{m}(\mathbf{x})
    \end{pmatrix} = \begin{pmatrix}
        O_m \\ M^{-1}(\mathbf{x}_1)
    \end{pmatrix}.
\end{equation}
\end{subequations}

To include the said control problem into the generalized integer-order Robust Control Framework, we consider a polytopic approximation of the system \eqref{eq:m_link_robot_ss}. 
The positions $\mathbf{q}$ and the velocities $\dot{\mathbf{q}}$ are bounded, forming a compact domain $\mathcal{D}_x$. Therefore, there exist matrices $A_i^{(j)}\in\mathbb{R}^{n_x\times n_x}$, $i=\overline{1,n_{A^{(j)}}}$, $j=\overline{0,m}$ such that:
\begin{equation}
    \label{eq:polytopic_cond}
    \frac{\partial g_j}{\partial \mathbf{x}} \in \text{Co}\left(\mathcal{A}^{(j)}\equiv \left\lbrace A^{(j)}_i,\ i=\overline{1,n_{A^{(j)}}} \right\rbrace\right), \ \forall \mathbf{x}\in \mathcal{D}_x,
\end{equation}
for each index $j=\overline{0,m}$.  

Now, the following polytopic linear differential inclusion (PLDI) will cover the behaviour of the given process: 
\begin{equation}
    \left(\Sigma_{\delta}\right): \ \ \begin{cases}
        \dot{\mathbf{x}}(t) = A(\delta)\mathbf{x}(t) + B(\delta)\mathbf{u}(t); \\
        \mathbf{y}(t) = \begin{pmatrix}
            I_{m} & O_{m}
        \end{pmatrix}\mathbf{x}(t),
    \end{cases}
\end{equation}
where $\mathbf{x}\in \mathbb{R}^{n_x}$, $\mathbf{u}\in \mathbb{R}^{n_u}$, $\delta \in U_{\delta} \subset \mathbb{R}^{n_{\delta}}$ is the uncertainty from the state and input matrices, and $U_{\delta}$ is closed and bounded. The PLDI should be characterized by the following $L$-vertex convex hull, according to \eqref{eq:polytopic_cond}:
\begin{equation}
\label{eq:convex_hull}
    \left\lbrace (A(\delta),B(\delta)) | \delta \in U_{\delta} \right\rbrace \subset \Omega \equiv \text{Co}\left\lbrace (A_i,B_i) , i=\overline{1,L} \right\rbrace.
\end{equation}

\textit{Assumption 1:} Each pair $(A(\delta),B(\delta))$ with $\delta \in U_{\delta}$ is stabilizable.

The nominal plant $G_n {=} (A(\mathbf{0}),B(\mathbf{0}),C,O)$ has as interface the set of control inputs $\mathbf{u}\in\mathbb{R}^{n_u}$ and the set of control outputs $\mathbf{y}\in\mathbb{R}^{n_y}$. The uncertain plant $G_{\bm{\Delta}} {\equiv} G$ presents an additional set of disturbance inputs $\mathbf{d}\in\mathbb{R}^{n_d}$ and an additional set of disturbance outputs $\mathbf{v}\in \mathbb{R}^{n_v}$ and it can be written as a function of the structured normalized uncertainty block $\bm{\Delta}$ of corresponding dimension with an adequate mapping: 
\begin{equation} \label{eq:T_delta_U}
    \mathcal{T} : \mathcal{G}^2 \to \mathcal{G}, \ G = \mathcal{T}\left(G_n,U\right), \hspace{1mm} \Delta \in \bm{\Delta}, \ \left\lVert \Delta \right\rVert_{\infty} \leq 1,
\end{equation}
having the transfer matrix $U$ partitioned in a similar manner with the structured uncertainty set $\bm{\Delta} = \{ \text{diag}(\delta_1I_{n_1},\dots,\delta_sI_{n_s}) \}$, 
where $\delta_iI_{n_i}$ is used to encompass the parametric uncertainties from $U_{\delta}$. 

\begin{figure}[!b]
    \centering
    \includegraphics[width=0.2\columnwidth]{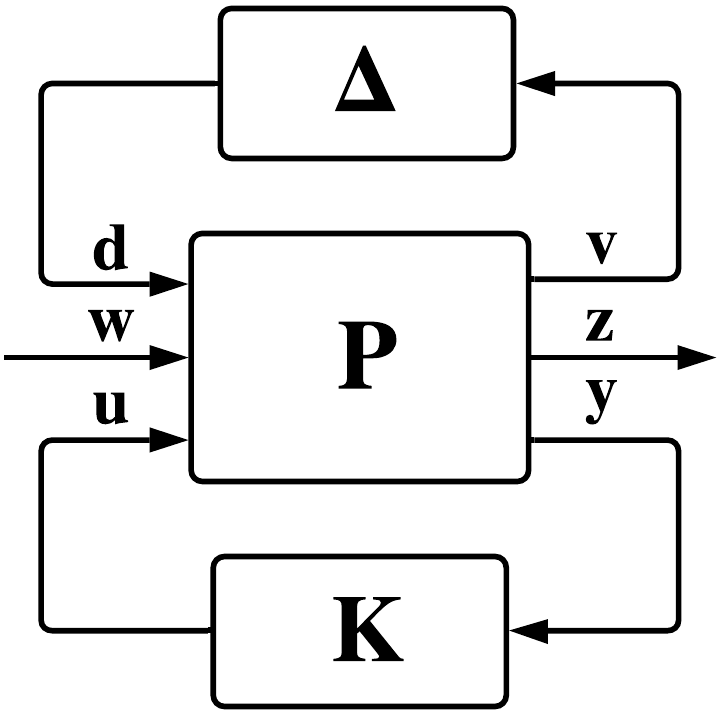}
    \caption{The generalized plant $P$ and its interconnections with the uncertainty block $\bm{\Delta}$ and the controller $K$.}
    \label{fig:P_Delta_K}
\end{figure}

To impose a set of performances, an input performance vector $\mathbf{w}\in \mathbb{R}^{n_w}$ and an output performance vector $\mathbf{z}\in \mathbb{R}^{n_z}$ should be considered, 
the augmented plant $P$ being written based on the uncertain plant $G$ through an adequate mapping $\mathcal{A}$ as $\mathcal{A} : \mathcal{G} \times \mathcal{G} \to \mathcal{G}, \ P = \mathcal{A}\left(G,W\right)$,
with a transfer matrix $W$ hosting the performance filters. The resulting continuous-time plant has the structure: 
\begin{equation}
    \label{eq:gen_plant}
 P : \begin{pmatrix}
\dot{{\mathbf{x}}}(t) \\ \hline 
\mathbf{v}(t) \\
\mathbf{z}(t) \\
\mathbf{y}(t)
\end{pmatrix} = \begin{pmatrix}
A & \vline & B_d & B_w & B_u \\
\hline
C_v & \vline & D_{vd} & D_{vw} & D_{vu} \\
C_z & \vline & D_{zd} & D_{zw} & D_{zu} \\
C_y & \vline & D_{yd} & D_{yw} & D_{yu}
\end{pmatrix} \begin{pmatrix}
\mathbf{x}(t) \\ \hline \mathbf{d}(t) \\ \mathbf{w}(t) \\ \mathbf{u}(t)
\end{pmatrix},
\end{equation}
where  $\mathbf{x}\in\mathbb{R}^{n_x}$ is the state vector. The interconnections between the plant $P$, the controller $K$, and the uncertainties block $\bm{\Delta}$ are illustrated in Figure \ref{fig:P_Delta_K}.


To compensate the uncertainty block, the structured singular value (SSV) of the closed-loop system given by the lower linear fractional transform interconnection between the plant $P$ and the controller $K$ according to the block $\bm{\Delta}$, denoted by  $\mu_{\bm{\Delta}}(\text{LLFT}(P,K))$, can be used:
\begin{equation*}
    \mu_{\bm{\Delta}}(\text{LLFT}(P,K)) = \sup_{\omega\in\mathbb{R}_+}\frac{1}{\displaystyle\min_{\Delta\in\bm{\Delta}}\{\overline{\sigma}(\Delta),\ \det(I{-}M_{\omega}\Delta){=}0\}},
\end{equation*}
where $M_{\omega} \vcentcolon = \text{LLFT}(P,K)(j\omega)$. The Main Loop Theorem 
states that a controller $K$ ensures robust stability and robust performance if $\displaystyle \mu_{\bm{\Delta}}(\text{LLFT}(P,K)) < 1$. Additionally, if a controller structure described by a family $\mathcal{K}$ is desired, the resulting optimization problem is:
\begin{equation}
\label{eq:mu_syn}
    \inf_{K\in\mathcal{K}} \mu_{\bm{\Delta}}(\text{LLFT}(P,K)).
\end{equation}

However, the computational problem regarding SSVs is non-deterministic polynomial-time (NP) hard. There are several approaches available to solve this problem, the most common being the so-called $D$--$K$ iteration approach, being based on an upper bound convexification procedure. The fixed-structure $\mu$-synthesis requires a possibility to solve a fixed-structure $\mathcal{H}_{\infty}$ control problem, which has been successfully solved \cite{Apkarian:17}.

\section{Numerical Example}

\begin{figure}
    \centering
    \includegraphics[width=\columnwidth]{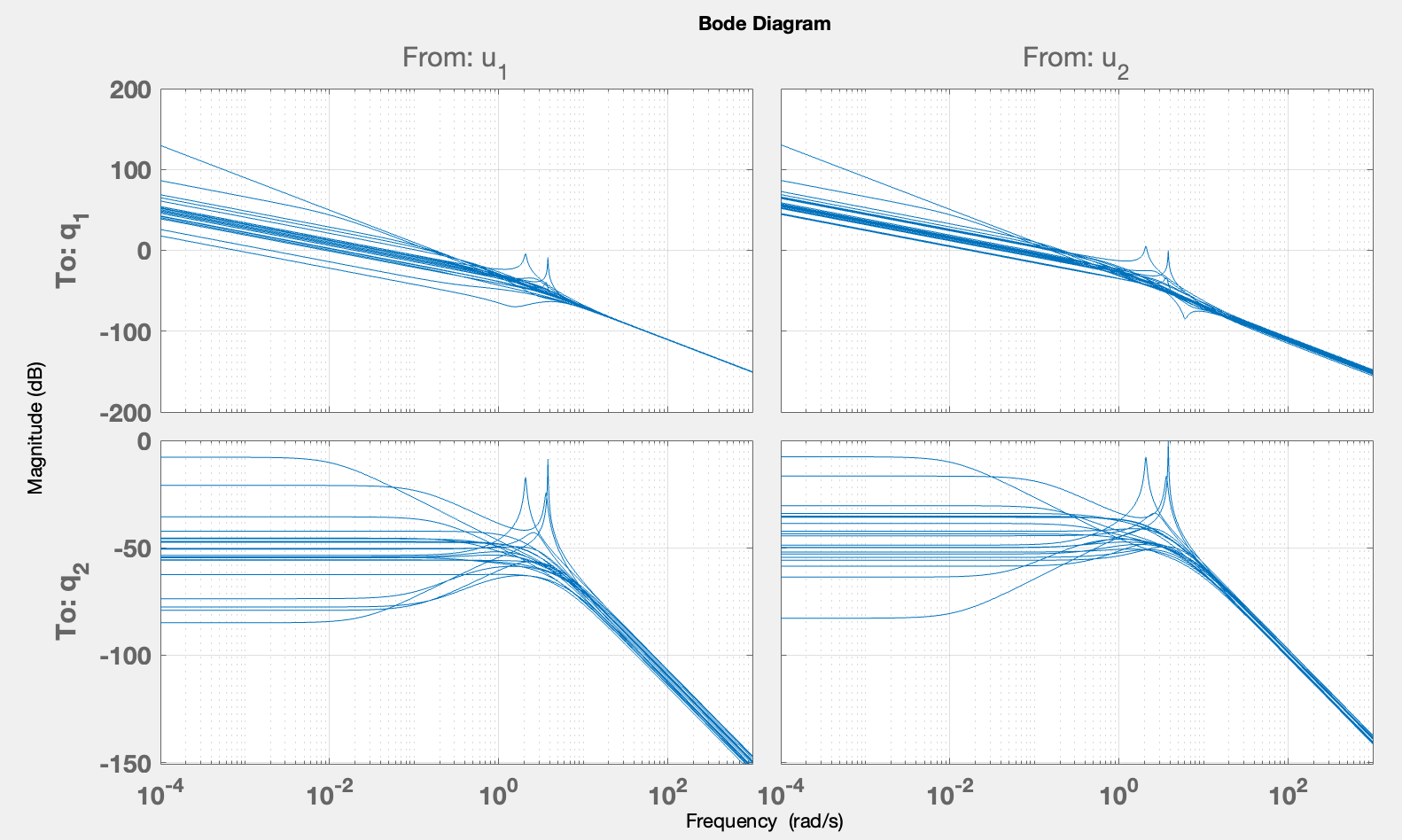}
    \caption{Magnitude Bode diagram of the linear polytopic differential inclusion $G_{\Delta}(s)$.}
    \label{fig:bode_process}
\end{figure}

In this section we present the proposed control methodology (for cancer treatment robots) on a 2R serial robot without friction (i.e. $D=0$). The functions involved in the state-space model \eqref{eq:m_link_robot_ss} are:
\begin{subequations}
\begin{equation}
    M(q_1,q_2) = \begin{pmatrix}
      a_1+2a_3\cos(q_2) & a_2+a_3\cos(q_2) \\ 
      a_2+a_3\cos(q_2) & a_2
    \end{pmatrix};
\end{equation}
\begin{equation}
    C(q_1,q_2,\dot{q}_1,\dot{q}_2) = a_3\sin(q_2)\begin{pmatrix}
      -\dot{q}_2 &  \dot{q}_1 +\dot{q}_2\\ 
      \dot{q}_1 & 0
    \end{pmatrix}.
\end{equation}
\end{subequations}
The gravity term will be also $g=0$, because the gravity acts along the Z axis.
The parameters of the above-mentioned model are: $a_1=48.125$, $a_2=13.125$, $a_3=6.25$. The state-space model can be viewed as the following linear polytopic differential inclusion:
\begin{equation}
G_{\Delta}(s): \ \ 
\begin{cases}
    \dot{\mathbf{x}} = \begin{pmatrix}
        0 & 0 & 1 & 0 \\
        0 & 0 & 0 & 1 \\
        0 & a_{32} & a_{33} & a_{34} \\
        0 & a_{42} & a_{43} & a_{44}
    \end{pmatrix}\mathbf{x} + \begin{pmatrix}
        0 & 0 \\
        0 & 0 \\
        b_{31} & b_{32} \\
        b_{41} & b_{42}
    \end{pmatrix}\mathbf{u}; \\
    \mathbf{y} = \begin{pmatrix}
        1 & 0 & 0 & 0 \\
        0 & 1 & 0 & 0
    \end{pmatrix}\mathbf{x},
\end{cases}
\end{equation}
where: $a_{32} \in [-19.127,19.6402]$, $a_{33} \in [-1.58,1.58]$, $a_{34} \in [-3.56,3.56]$, $a_{42} \in [-13.9637,28.2362]$, $a_{43} \in [-5.42,5.42]$, $a_{44} \in [-3.95,3.95]$, $b_{31} \in [0.0286,0.0312]$, $b_{32} \in [-0.0461,-0.0164]$, $b_{41} \in [-0.0461,-0.0164]$, $b_{42} \in [0.0848,0.144]$. 
The magnitude Bode diagram of the given process is described in Figure \ref{fig:bode_process}. Such magnitude Bode diagrams are used to represent the frequency response of linear systems. In Control Engineering, these are great substitutes for studying both stability and performance aspects simultaneously.

To impose a set of desired performances which should be met by each process form $G_{\Delta}(s)$, we consider an augmentation using the sensitivity and the complementary sensitivity functions. We consider the following shapes of the weighting functions used for augmenting the plant \cite{Skogestad:05}: 
\begin{equation}
    W_S(s) = \frac{1/M_S \cdot s + \omega_B}{s + \omega_B\cdot A} \ \ \text{and} \ \ W_T(s) = \frac{s+\omega_{BT}}{A_Ts+\omega_{BT}\cdot M_T}.
\end{equation}

The rise time can be imposed using the minimum allowed bandwidth $\omega_B$, the maximum overshoot can be imposed via the maximum amplitude of the sensitivity function $M_S$, while the maximum allowed steady-state error is imposed by the magnitude at low frequencies of the sensitivity function $A_S$. Similarly, for the complementary sensitivity we consider $\omega_{BT} > 10 \omega_B$, $M_T \approx M_S$ and $A_T \approx A_S$. 

For the proposed use-case we consider the following hyperparameters, one for each input-output pair: $\omega_{B,1} = 0.5[rad/s]$, $\omega_{B,2}=0.1[rad/s]$, $M_{S,1}=2$, $M_{S,2}=3$, $A_{S,1}=10^{-2}$, and $A_{S,2}=2\cdot 10^{-2}$. As such, for complementary sensitivity we have: $\omega_{BT,1}=10[rad/s]$, $\omega_{BT,2}=12[rad/s]$, $M_{T,1}=2.1$, $M_{T,2}=3$, $A_{T,1}=A_{T,2}=10^{-2}$. The resulting weighting functions are given by the following block-diagonal transfer matrices: 
\begin{equation}
    W_{S}(s) = \begin{pmatrix}
        \frac{0.5s+0.5}{s+0.005} & 0 \\
        0 & \frac{0.3333s+0.1}{s+0.002}
    \end{pmatrix} \ \ \text{and} \ \ W_T(s) = \begin{pmatrix}
        \frac{s+10}{0.01s+21} & 0 \\ 0 & \frac{s+12}{0.01s+36}
    \end{pmatrix}.
\end{equation}

\begin{figure}
    \centering
 \includegraphics[width=\columnwidth]{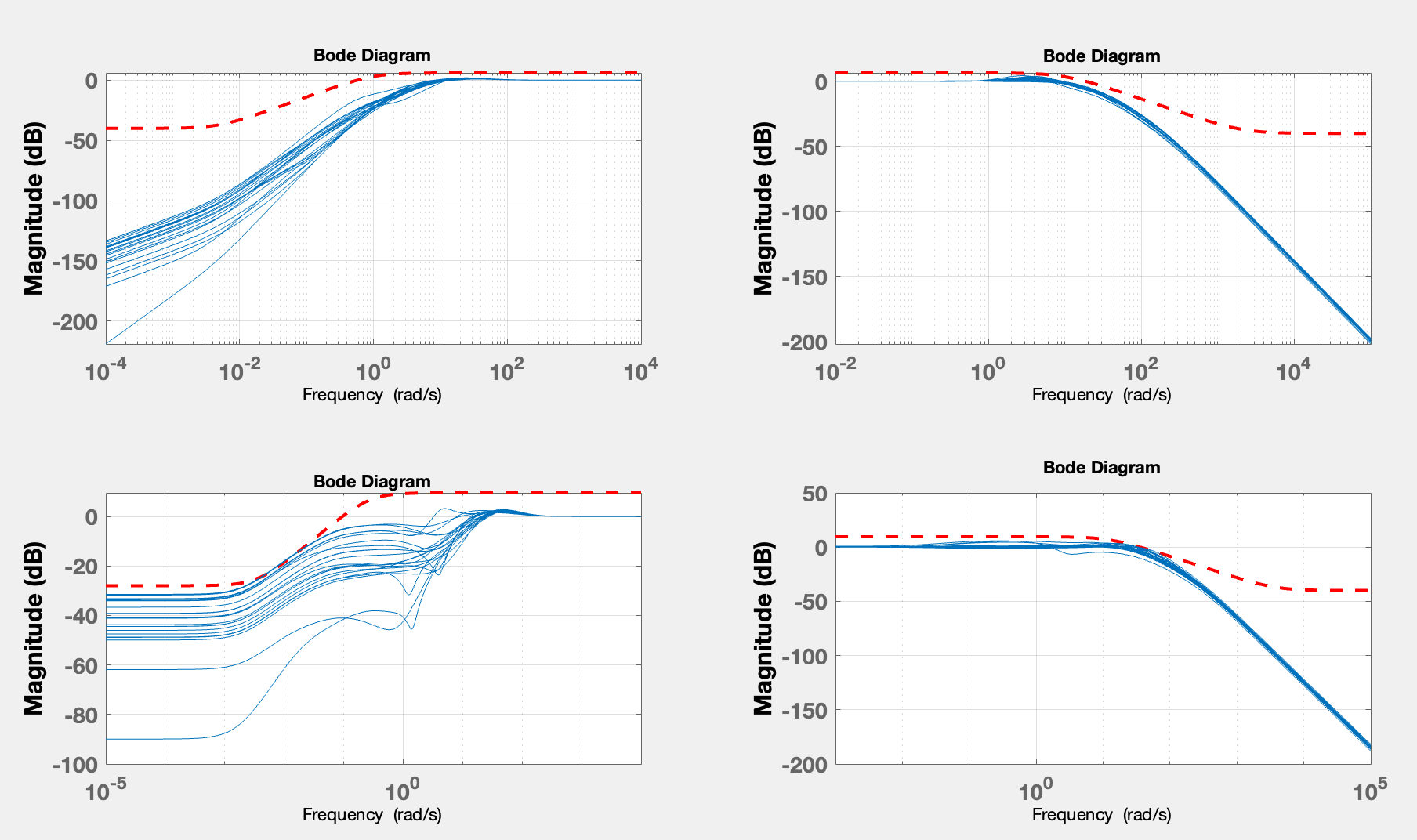}
    \caption{The evolution of the sensitivity and the complementary sensitivity functions for both nominal model and $20$ Monte Carlo simulations, along with the imposed shapes via the augmentation process (red line).}
    \label{fig:S_T}
\end{figure}

Initially, the unstructured case has been considered. Using the \texttt{musyn} routine from MATLAB, a controller $K$ which ensures both robust stability and robust performance is obtained after $22$ $D$--$K$ iterations. The mathematical guarantee of the robustness is given by the upper bound of the structured singular value of the closed-loop system according to the uncertainty set $\mu_{\bm{\Delta}(\text{LLFT}(P,K))}\leq 0.9998 < 1$. An illustrative plot which certifies that the controller ensures the robustness properties is presented in Figure \ref{fig:S_T}. 

However, the main drawback of the unstructured approach consists in the high order of the resulting controller, which could represent a major issue for the numerical implementation process \cite{suscaCDC2022}. In the above illustrated case the regulator is of order $132$. A solution is to impose a fixed-structure controller: 
\begin{equation}
    K(s) = \begin{pmatrix}
        K_{11}(s) & K_{12}(s) \\ K_{21}(s) & K_{22}(s)
    \end{pmatrix} \in \mathcal{K}.
\end{equation}
For the purpose of this paper we impose each component of $K(s)$ to be of third order. Solving the fixed-structure $\mu$-synthesis control problem using \texttt{musyn} routine, a robust controller has been obtained in $25$ $D$--$K$ iterations, with an upper bound of the structured singular value $\mu_{\bm{\Delta}}(\text{LLFT}(P,K))\leq 0.9999 < 1$.
The resulting controller is: 
\begin{equation}
    K^{\star}(s) = \begin{pmatrix}
        \frac{1.07\cdot 10^5 s^2 + 1.392\cdot 10^5 s + 44157}{0.02 s^3 + 4 s^2 + 200 s + 1} & \frac{264150 s^2 + 5.362e05 s + 15849}{ 0.08333 s^3 + 13.33 s^2 + 500 s + 1} \\ 
        \frac{ 3.421\cdot 10^4 s^2 + 4.447\cdot 10^4 s + 14454}{ 0.01422 s^3 + 3.556 s^2 + 222.2 s + 1} & \frac{ 1.172\cdot 10^5 s^2 + 4.728\cdot 10^5 s + 16406}{0.02319 s^3 + 6.494 s^2 + 454.6 s + 1}
    \end{pmatrix}.
\end{equation}

\begin{figure}
    \centering
    \includegraphics[width=\columnwidth]{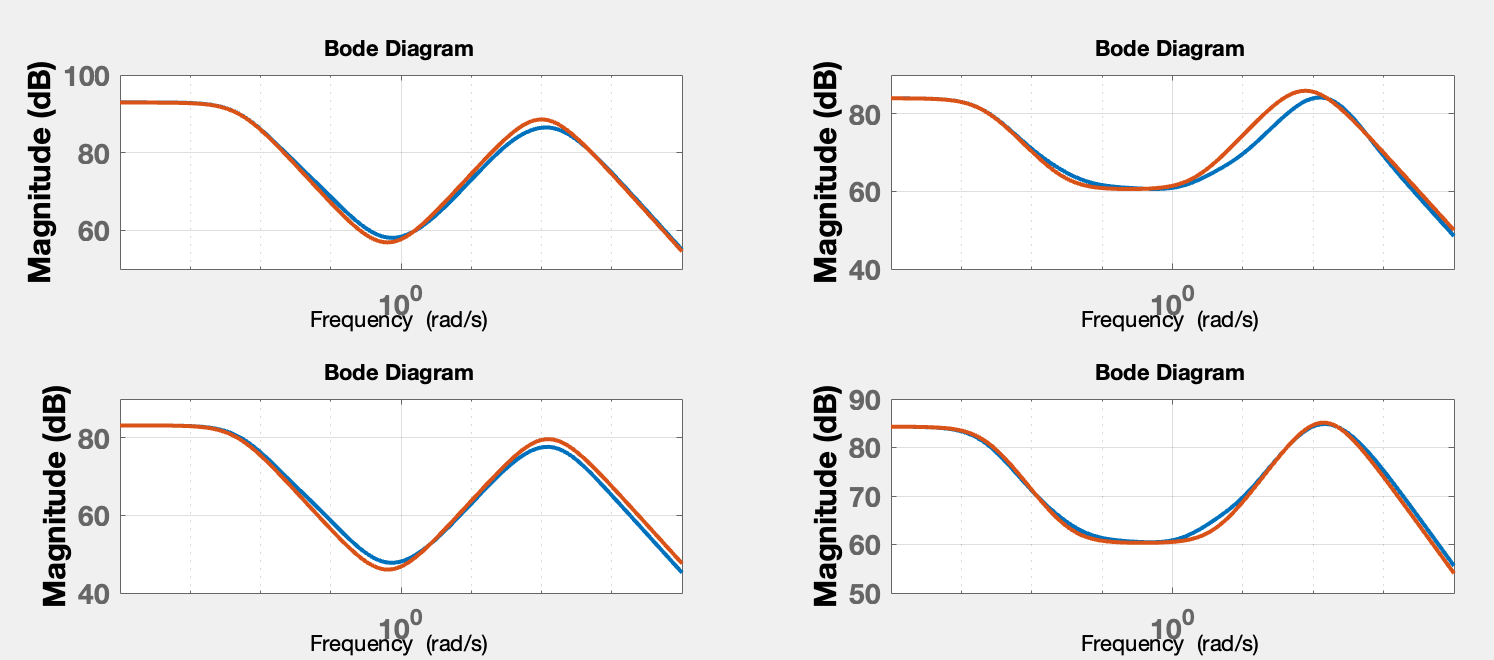}
    \caption{Bode magnitude representations of the robust high-order controller (blue) and of the fixed-structure robust controller (orange).}
    \label{fig:K_vs_K_star}
\end{figure}

A comparison between the initial high-order robust controller and the fixed-structure robust controller is depicted in Figure \ref{fig:K_vs_K_star}. As noticed, the differences are small, with a significant benefit regarding the implementation. Moreover, there is no need of an integrative effect, so the controller is stable. It allows one to perform a discretization and a quantization analysis in a straightforward manner.  

\section{Conclusions}
Cancer treatment robotic systems require adequate control techniques to achieve patient safety in personalized therapies. The paper proposes a method to include the nonlinear model of a general m-link serial surgical robot into the generalized Robust Control Framework. Even if the linear polytopic differential inclusion could be conservative, a mathematical guarantee for an imposed set of performances is still fulfilled. The great advantage of the proposed method against the available solutions, such as robust passivity-based control, is the ability to reduce the control structure to a single layer. Numerical examples showed an incipient implementation issue regarding the order of the approximation, and an fixed-structure robust controller has been computed, without losing the robustness guarantee. 

Further work is required to implement the proposed control techniques into existing medical robots and novel cancer treatment robots. Another research direction will be to extend the proposed control structure for parallel robots, which are usually described using input-affine descriptor nonlinear systems, presenting additional algebraic constraints.

\subsubsection{Acknowledgements} 
This research was supported by the project New smart and adaptive
robotics solutions for personalized minimally invasive surgery in
cancer treatment - ATHENA, funded by European Union – NextGenerationEU
and Romanian Government, under National Recovery and Resilience Plan
for Romania, contract no. 760072/23.05.2023, code CF 116/15.11.2022,
through the Romanian Ministry of Research, Innovation and
Digitalization, within Component 9, investment I8.

%
%
%
%

\end{document}